\documentclass[runningheads]{llncs}
\usepackage{graphicx}
\begin{document}
\title{Rethinking Voxelization and Classification for 3D Object Detection\thanks{Supported in part of method development by a grant from the Russian Science Foundation No. 21-71-00131 and in part of experimental study by a grant for research centers in the field of artificial intelligence, provided by the Analytical Center for the Government of the Russian Federation in accordance with the subsidy agreement (agreement identifier 000000D730321P5Q0002) and the agreement with the Moscow Institute of Physics and Technology dated November 1, 2021 No. 70-2021-00138.}}
%
%
\author{Youshaa Murhij\inst{1}\orcidID{0000-0003-2036-4023} \and
Alexander Golodkov\inst{1} \and
Dmitry Yudin\inst{1,2}\orcidID{0000-0002-1407-2633}}
%
%
\institute{Moscow Institute of Physics and Technology, Dolgoprudny, Moscow Region, Russia
\\
\and
Artificial Intelligence Research Institute, Moscow, Russia\\
\email{\{yosha.morheg,golodkov.ao\}@phystech.edu, yudin.da@mipt.ru}}
\maketitle              
\begin{abstract}
The main challenge in 3D object detection from LiDAR point clouds is achieving real-time performance without affecting the reliability of the network. In other words, the detecting network must be confident enough about its predictions.
In this paper, we present a solution to improve network inference speed and precision at the same time by implementing a fast dynamic voxelizer that works on fast pillar-based models in the same way a voxelizer works on slow voxel-based models. In addition, we propose a lightweight detection sub-head model for classifying predicted objects and filter out false detected objects that significantly improves model precision in a negligible time and computing cost. The developed code is publicly available at:

\url{https://github.com/YoushaaMurhij/RVCDet}.

\end{abstract}
\section{Introduction}
Current LiDAR-based 3D object detection approaches follow a standard scheme in their pipelines. Most of 3D detection pipelines consists of reading module that prepares point clouds for voxelization stage that converts raw points into a fixed size 2D or 3D grid which can be fed to a detection neural network. Most common grid formats use voxels or pillars in this stage. Compared to other methods like 2D projected images and raw Lidar point, voxel representation can be processed efficiently using 3D sparse convolution \cite{chen2022focal} and preserve approximately similar information to raw point cloud and make feature sparsity learnable with position-wise importance prediction. A Pillar represents multiple voxels which are vertically stacked and treated as a one tall voxel. Voxels are generally used in 3D backbones such as VoxelNet \cite{zhou2018voxelnet}, while pillars are used in 2D backbones such as PointPillars \cite{lang2019pointpillars}.
In this paper, we are going to discuss the differences between the two representations and introduce a new voxelization approach to benefit from voxel features in pillars representations by implementing a fast dynamic voxel encoder for 2D backbones like PointPillars.

Current 3D detection neural nets suffer from noisy outputs which can be seen as false positive objects in the network predictions.  This problem can be reduced by filtering the network output based on each object score (How much the network is confident that this object is corresponding to a certain class). But this will reduce the network precision as it could filter out a true positive object with low confidence score.
To address this problem,  we present an auxiliary module that can be merged to the detection head in the network and works as classification sub-head. Classification sub-head learns to distinguish true and false objects base on further processing intermediate features generated by neck module in the detection pipeline.

\section{Related Work}
Different forms of point cloud representation have been explored in the context of 3D object detection. The main idea is to form a structured representation where standard
convolution operation can be applied. 

Existing representations are mainly divided into two types: 3D voxel grids and 2D projections. A 3D voxel grid transforms the point cloud into a regularly spaced 3D grid, where each voxel cell can contain a scalar value (e.g., occupancy) or vector data (e.g., hand-crafted statistics computed from the points within that voxel cell). 3D convolution is typically applied to extract high-order representation from the voxel grid \cite{engelcke2017vote3deep}. However, since point clouds are sparse by nature, the voxel grid is very sparse and therefore a large proportion of computation is redundant and unnecessary. As a result, typical systems \cite{engelcke2017vote3deep}, \cite{li2016vehicle} only run at 1-2 FPS.

Several Point cloud based 3D object detection methods utilize a voxel grid representation. \cite{engelcke2017vote3deep} encode each non-empty voxel with 6 statistical quantities that are derived from all the points contained within the voxel. \cite{schwarz2022voxgraf} fuses multiple local statistics to represent each voxel. \cite{song2016deep} computes the truncated signed distance on the voxel grid. \cite{li20173d} uses binary encoding for the 3D voxel grid. \cite{chen2017multi} introduces a multi-view representation for a LiDAR point cloud by computing a multi-channel feature map in the bird's eye view and the cylindral coordinates in the frontal view. Several other studies project point clouds onto a perspective view and then use image-based feature encoding schemes \cite{premebida2014pedestrian}, \cite{li2016vehicle}. While Center-based 3D object detecion and tracking method \cite{yin2021center}, \cite{youshaa2022fmfnet} are designed based on \cite{duan2019centernet}, \cite{zhu2019classbalanced} to represent, detect, and track 3D objects as points. CenterPoint framework, first detects centers of objects using a keypoint detector and regresses to other attributes, including 3D size, 3D orientation, and velocity. In a second stage, CenterPoint refines these estimates using additional point features on the object. In CenterPoint, 3D object tracking simplifies to greedy closest-point matching.
An attempt to synergize the birds-eye view and the perspective view was done in \cite{zhou2019endtoend}  through a novel end-to-end multiview fusion (MVF) algorithm, which can learn to utilize the complementary information from both. 

A pipeline for multi-label object recognition, detection and semantic segmentation was introduced in \cite{Ge_2018_CVPR} benefiting from classification approach on 2D image data. In this pipeline, They tried to obtain intermediate object localization and pixel labeling results for the training data, and then use such results to train task-specific deep networks in a fully supervised manner. The entire process consists of four stages, including object localization in the training images, filtering and fusing object instances, pixel labeling for the training images, and task-specific network training. To obtain clean object instances in the training images, they proposed an algorithm for filtering, fusing and classifying predicted object instances.

In our work, we will also show that adding an auxiliary classifier to 3D detection head will improve the model precision as this sub-head will learn not from positive ground truth examples but from negative examples (false positive), too. Serveral approaches are available to reduce the number of false positives in 3D object detection. One of them is adding a point cloud classifier. As an input, the classifier takes the points clouds which are included in the predicted 3D boxes by the detector, and returns which class the given box belongs to: true or false predictions. The detector predictions contain many false positives, and, accordingly, by reducing their number, one can significantly improve the accuracy of the approach used. At the moment, there are three main approaches for point cloud classification. 

Algorithms based on voxels \cite{octnet}, which represent the classified point cloud as a set of voxels. Further, 3D grid convolutions are usually used for classification. The downside of this approach is that important information about the geometry of points can be lost, while voxels complicate calculations and such algorithms require more memory. Projective approaches \cite{projective_conv}. These algorithms usually classify not the point cloud itself, but its projections on planes, and convolutional networks are usually used for classification. The disadvantage of such approaches is that data on the spatial relative position of the points is lost, which can be critical if there are few points in the classified cloud. Algorithms that process points directly \cite{pointnet}. Usually these are algorithms that apply convolutions to points directly and take into account the density of points. Such algorithms are invariant with respect to spatial displacements of objects. All this makes this type of algorithm the most interesting for use in the problem of classifying lidar point clouds. 

PAConv \cite{paconv} from the third group, and one of the most effective classification models was trained, and evaluated on ModelNet40 \cite{modelnet} dataset. PAConv shows a prediction accuracy of 93.9 on objects with 1024 points. But, this approach was not used in conjunction with our detection pipeline due to the large inference time, which made the pipeline not real-time. An alternative to such a classifier is a lighter network, which is an MLP, possibly with the addition of a small number of convolutional layers. It is proposed to classify intermediate tensors in the sub-heads of the detector, containing information about the detected boxes. This topic will be discussed in this article in details.

\section{Method}
Our 3D detection pipeline consists of a fast dynamic voxelizer (FDV), where we implemented our fast voxelization method for pillar-based models based on scatter operations, a 2D backbone for real-time performance (RV Backbone), which is adapted to take multiple-channel FDV features as an input, regional proposal network (RPN module) as neck for feature map generation and a Center-based detection head to predict 3D bounding boxes in the scene with our additional sub-head classifier to filter out false detected objects.
Next, in this section, we discuss the proposed FDV voxelizer in \ref{sssec:voxel}, our adapted RV backbone that accepts FDV output in \ref{sssec:backbone} and our proposed classifier sub-head in \ref{sssec:classifier}.
\subsection{Fast Dynamic Voxelizer}  \label{sssec:voxel}
We have implemented a fast dynamic voxelizer (FDV) for PointPillars model, which works similarly to the dynamic 3D voxelizer for the VoxelNet model. The proposed voxelizer is based on scatter operations on sparse point clouds. FDV voxelizer runs with $O(N)$ time complexity faster than most current similar approaches.
FDV implementation does not require to sample a predefined number of points per voxel. This means that every point can be used by the model, minimizing information loss. Furthermore, no need to pad voxels to a predefined size, even when they have significantly fewer points. This can greatly reduce the extra space and compute overhead compared to regular methods (hard voxelization), especially at longer ranges where the point cloud becomes very sparse.

\begin{figure*}[ht]
\begin{center}
\includegraphics[width=0.5\linewidth]{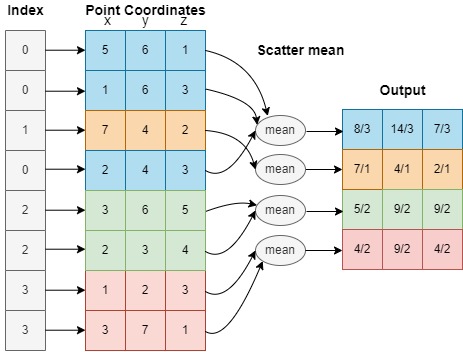}
\end{center}
   \caption{Scatter Mean operation over 2D/3D voxel grid}
\label{fig:scatter_mean}
\end{figure*}

As an input, we feed a point cloud in the format: $Batch\_id,x,y,z$. We determine the size of the grid by the voxel size and the range of the input point cloud after filtering not in-range points. $gridsize_i = cloudrange_i / voxelsize_i$, where $i$ represents size along $x, y, z$ axis. After that, we determine the coordinates of the voxels on the grid from the point cloud range and remove the redundant points (this may happen in the center of the coordinates). Next, we calculate the mean of all points in each non-empty pillar using the scatter mean method and find the distance $x, y$ and $z$ from the center of the voxel to the calculated mean point. 
Scatter mean averages all values from the $src$ tensor into $out$ at the indices specified in the index tensor along a given axis dimension as seen in Fig. \ref{fig:scatter_mean}
For one-dimensional tensors, the operation computes: 
\begin{equation}
    out_i = out_i + \frac{1}{N_i}.\sum_{j}{src_j}
\end{equation}
where $\sum_{j}$ is over $j$ such that $index j = i.N_i$ indicates referencing $i$.

Finally, we combine all the features together and pass them to the RV backbone module. The features include $x_{pt}, y_{pt}, z_{pt}$ for each point in the voxel, $x_{center}$, $y_{center}$, $z_{center}$ the distance for each point to the voxel center and $x_{mean}$, $y_{mean}$, $z_{mean}$ the distance from each point to the mean point. 

\subsection{RV Backbone module} \label{sssec:backbone}
After concatenating all the features from FDV module, we feed them to our proposed RV backbone.
RV backbone takes advantage from the gathered scatter data as it consists of multiple modified Pillar Feature Nets (PFN) for further feature extraction. The main difference between our RV backbone and orignal FPN \cite{lang2019pointpillars} is that RV backbone accepts input features from multiple channels in a similar way voxel feature encoder (VFE) \cite{zhou2018voxelnet} works in voxel-based models. In our implementation, RV backbone includes two Pillar-like feature nets with additional scatter layer to get the features maximum values (scatter-max) from each sub-module. Next, we re-concatenate the new features, again calculate their final scatter-max and rearrange them to generate initial feature maps which we feed to RPN module.

\subsection{Classification sub-head module} \label{sssec:classifier}
Mean average precision metric does not provide rich information about the model performance in terms of true and false positive detections. We noticed that even if a certain model gives a high mAP metric, it still detects enough amount of false positive objects (FP), which is a major issue in self-driving real-world scenarios. To tackle this problem, we propose an additional refining module in our detection pipeline to classify and filter FP predictions based on their features though a classification sub-head in our Center-based detection head by additional processing of the corresponding middle-features learned in the neck module.  

The main reason to add such a sub-head in our model structure is that after voxelizing the input point cloud into 2d grid and generating spatial features though a scatter backbone, we feed these middle-features into a regional proposal network (RPN) that calculate the final features and provide a set of heat-maps (HM) that can be directly used to regress all the 3D attributes of the predicted output (position, scale, orientation, class and score). So, these temporary features are sufficient to achieve this task by further prepossessing and training. In addition to regressing to all these attributes, we add NN Sub-Head module that takes the crops of HM tensors corresponding to the predicted objects and refine them to False/True positive predictions depending on the number of classes we train on.

\begin{figure*}[ht]
\begin{center}
\includegraphics[width=1.0\linewidth]{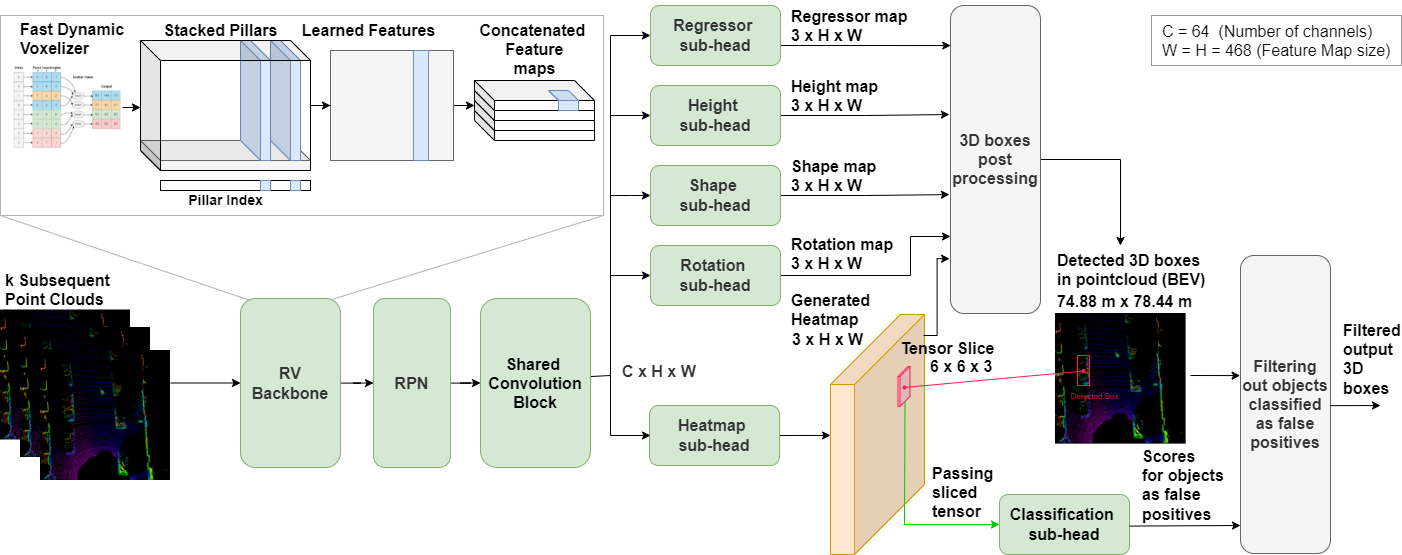}
\end{center}
  \caption{Scheme of our 3D detection pipeline named as RVDet including classification sub-head}
\label{fig:pipeline}
\end{figure*}

The proposed method pipeline is illustrated in Fig \ref{fig:pipeline}.

As seen in Figure \ref{fig:pipeline}, we are interested in HM sub-head as it represents the Gaussian distribution map and contains all the required information to locate the objects in the scene. According to the proposed scheme the classifier takes the predicted output of the detection pipeline as an input in addition to middle features (HM) generated in the RPN module. Our sub-head classifier crops the input HM tensors to multiple small windows including the predicted objects based on their predicted coordinates in 3D space and maps them to their 2D plane coordinates in the feature map so, it can focus only on the objects of interest in the generated map.

Next, our classification module refines these predictions based on their initialized predicted classes into two categories: true positive class and false negative class. Where, Class could be either {Vehicle, Pedestrian or Cyclist} in Waymo or KITTI datasets.

Based on the predicted TP/FP classes it refines the initial pipeline predictions and returns the final predictions. Figure \ref{fig:pipeline} shows more detailed scheme illustrating the tensor cropping process and classification.

Regarding the classifier's architecture, we have implemented several models based on multi-layer perceptrons (MLP) and convolutional layers (Conv). Our motivation while choosing the architecture was to provide simple and lightweight module that can achieve this task considering real-time performance of the pipeline.

To train the classification sub-head, we prepared a custom dataset for this purpose by running our detection model in inference mode on Waymo and KITTI datasets. As an input data, we stored the feature heat map generated after neck module. We classified the network predictions into: False Vehicle, False Pedestrian, False Cyclist according to the ground-true data from the original dataset. During training, we used cross entropy as loss function, Adam as optimizer. 

\subsection{Datasets}
We trained and evaluated our pipeline on Waymo open dataset for perception \cite{sun2020scalability} and on KITTI vision dataset \cite{Geiger2012CVPR}. The Waymo perception dataset \cite{sun2020scalability} contains 1,950 segments of 20s each, collected at 10Hz (390,000 frames) in diverse geographies and conditions. The Sensor data includes: 1 mid-range lidar, 4 short-range lidars, 5 cameras (front and sides), Synchronized lidar and camera data, Lidar to camera projections, Sensor calibrations and vehicle poses. While Labeled data includes: Labels for 4 object classes - Vehicles, Pedestrians, Cyclists, Signs, High-quality labels for lidar data in 1,200 segments, 12.6M 3D bounding box labels with tracking IDs on lidar data, High-quality labels for camera data in 1,000 segments, 11.8M 2D bounding box labels with tracking IDs on camera data.

The Kitti 3D object detection benchmark \cite{Geiger2012CVPR} consists of 7481 training images and 7518 test images as well as the corresponding point clouds, comprising a total of 80.256 labeled objects. 

\begin{table}[h]
\caption{Dynamic voxelizer impact on 3D detection metrics on Waymo test dataset}
\begin{center}
\scriptsize
\begin{tabular}{lrrrrrrrr}
\hline
\textbf{Model} & \textbf{Range} & \textbf{mAP/L1} & \textbf{mAPH/L1} & \textbf{mAP/L2} &  \textbf{mAPH/L2} \\
\hline\hline
PointPillars \cite{lang2019pointpillars}
        & PerType	    & 0.4406	& 0.3985	& 0.3933	& 0.3552 \\
        & [0, 30)	    & 0.5322	& 0.4901	& 0.5201	& 0.4792 \\
        & [30, 50)	    & 0.4225	& 0.3792	& 0.3859	& 0.3458 \\
        & [50, +inf)	& 0.3010	& 0.2594	& 0.2376	& 0.2038 \\
\hline
SECOND- \cite{zhou2018voxelnet} 
         & PerType	    & 0.4364	& 0.3943	& 0.3914	& 0.3530 \\
VoxelNet & [0, 30)	    & 0.5291	& 0.4865	& 0.5134	& 0.4721 \\
         & [30, 50)	    & 0.4211	& 0.3771	& 0.3846	& 0.3440 \\
         & [50, +inf)	& 0.2981	& 0.2571	& 0.2366	& 0.2030 \\
\hline
CIA-SSD \cite{Zheng2021CIASSDCI} 
        & PerType	    & 0.6148	& 0.5794	& 0.5605	& 0.5281 \\
        & [0, 30)	    & 0.8002	& 0.7630	& 0.7873	& 0.7509 \\
        & [30, 50)	    & 0.5946	& 0.5549	& 0.5520	& 0.5153 \\
        & [50, +inf)	& 0.3085	& 0.2763	& 0.2482	& 0.2222 \\
\hline
QuickDet    
        & PerType	    & 0.7055	& 0.6489	& 0.6510	& 0.5985 \\
        & [0, 30)	    & 0.7995	& 0.7418	& 0.7841	& 0.7280 \\
        & [30, 50)	    & 0.6905	& 0.6385	& 0.6457	& 0.5969 \\
        & [50, +inf)	& 0.5484	& 0.4863	& 0.4620	& 0.4089 \\
\hline
CenterPoint-
        & PerType	    & 0.7093	& 0.6877	& 0.6608	& 0.6404 \\
PointPillars \cite{yin2021center}
        & [0, 30)	    & 0.7766	& 0.7501	& 0.7627	& 0.7367 \\
        & [30, 50)	    & 0.7095	& 0.6892	& 0.6692	& 0.6497 \\
        & [50, +inf)	& 0.5858	& 0.5698	& 0.5042	& 0.4899 \\
\hline
\bf{RVDet (ours*)} 
        & \bf{PerType}  &\bf{0.7488}&\bf{0.7280}&\bf{0.6986}&\bf{0.6789} \\
        & [0, 30)	    & 0.8503	& 0.8268	& 0.8357	& 0.8127 \\
        & [30, 50)      & 0.7195	& 0.7001	& 0.6787	& 0.6602 \\
        & [50, +inf)	& 0.5873	& 0.5706	& 0.5061	& 0.4911 \\
\hline\hline
CenterPoint- 
            & PerType    &	0.7871 &	0.7718 &	0.7338 &	0.7193 \\
VoxelNet \cite{yin2021center} 
            & [0, 30)    &	0.8766 &	0.8616 &	0.8621 &	0.8474 \\
            & [30, 50)   &	0.7643 &	0.7492 &	0.7199 &	0.7055 \\
            & [50, +inf) &  0.6404 &	0.6248 &	0.5522 &	0.5382 \\
\hline
CenterPoint++   & PerType	    & 0.7941	& 0.7796	& 0.7422	& 0.7282 \\
VoxelNet \cite{yin2021center}
                & [0, 30)	    & 0.8714	& 0.8568	& 0.8566	& 0.8422 \\
                & [30, 50)	    & 0.7743	& 0.7605	& 0.7322	& 0.7189 \\
                & [50, +inf)	& 0.6610	& 0.6469	& 0.5713	& 0.5586 \\
\hline
\end{tabular}
\end{center}
\label{tab:dyn_voxel_waymo_test}
\end{table}
\begin{table}[h]
\caption{Dynamic voxelizer main results on 3D detection KITTI test dataset}
\begin{center}
\scriptsize

\begin{tabular}{l | l | c | c | c | c | c | c }
\hline
 \multicolumn{2}{c}{ }  & \multicolumn{2}{|c|}{\bf Car} & \multicolumn{2}{c|}{\bf Pedestrian} & \multicolumn{2}{c}{ \bf Cyclist}  \\
\hline
\bf Model & \bf Difficulty & \bf 3D Det & \bf BEV & \bf 3D Det & \bf BEV & \bf 3D Det &  \bf BEV \\
\hline \hline
CenterPoint-
                & Easy     & 82.58 & 90.52 & 41.31 & 48.21 & 68.78 & 72.16 \\
PointPillars \cite{yin2021center}
                & Moderate & 72.71 & 86.60 & 35.32 & 41.53 & 52.52 & 57.43 \\
                & Hard     & 67.92 & 82.69 & 33.35 & 39.82 & 46.77 & 51.28 \\
\hline
\bf RVDet (ours)
                & Easy     & 82.37     &  89.90 & \bf 45.93  & \bf 52.91 & \bf 69.50  & \bf 76.11 \\
                & Moderate & \bf 73.50 &  86.21 & \bf 38.54  & \bf 45.18 & 51.78      & \bf 58.40 \\
                & Hard     & \bf 68.79 &  82.63 & \bf 36.26  & \bf 42.99 & 46.05      & \bf 51.93 \\
\hline\hline
CenterPoint-
                & Easy     & 81.17 & 88.47 & 47.25 & 51.76 & 72.16 & 76.38 \\
VoxelNet \cite{yin2021center}
                & Moderate & 73.96 & 85.05 & 39.28 & 44.08 & 56.67 & 61.25 \\
                & Hard     & 69.48 & 81.19 & 36.78 & 41.80 & 50.60 & 54.68 \\
\hline
PV-RCNN \cite{shi2020pv} 
                & Easy     & 90.25 & 94.98 & 52.17 & 59.86 & 78.60 & 82.49 \\
                & Moderate & 81.43 & 90.65 & 43.29 & 50.57 & 63.71 & 68.89 \\
                & Hard     & 76.82 & 86.14 & 40.29 & 46.74 & 57.65 & 62.41 \\
\hline
\end{tabular}
\end{center}
\label{tab:dyn_cp_pp_kitti_test_}
\end{table}

\section{Experimental Results}
\begin{figure*}[ht]
\begin{center}
\includegraphics[width=0.8\linewidth]{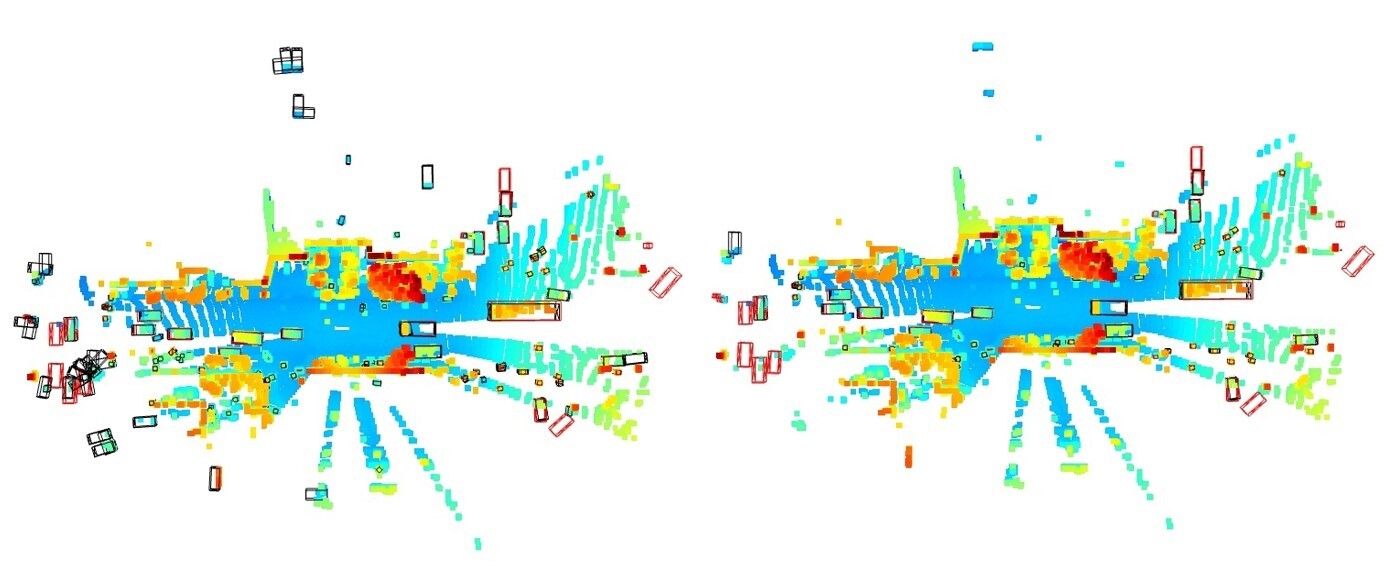}
\end{center}
  \caption{Detection pipeline performance example on Waymo before (left) and after (right) adding classification subhead}
\label{fig:pc_example}
\end{figure*}

We mainly tested our voxelizer and classifier on two popular datasets for 3D object detection: Waymo Open Dataset \cite{sun2020scalability} and KITTI dataset \cite{Geiger2012CVPR}.
We compared our pipeline performance with other current open real-time approaches on Waymo leaderboard. Our implementation with PointPillars backbone added +3\% $mAPHL2$ on Waymo testset outperforming all other PointPillars-based models while running +18FPS on Tesla-V100. This implementation is now comparable with Heavy VoxelNet-based models that run under 11FPS on Tesla-V100. Table \ref{tab:dyn_voxel_waymo_test} shows our pipeline mean average precision metrics with/without heading (both Level 1 and 2) on Waymo testset compared to real-time PointPillars models and VoxelNet models.
Table \ref{tab:dyn_cp_pp_kitti_test_} shows the performance of our pipeline on KITTI testset. Our RVDet model including FDV and RV backbone achieved +1 mAP on Car class, +2.4 mAP on Pedestrian class and +1.4 mAP on Cyclist class. Again on KITTI testset our real-time PointPillars implementation is comparable with CenterPoint VoxelNet method.
Figure \ref{fig:pc_example} shows detection pipeline performance example on Waymo before and after adding an classifier subhead. Red boxes refer to ground true data while black ones are network predictions. We show more detailed metrics related to classifier's performance in section \ref{sssec:ablation}.

\begin{table}[h]
\caption{Dynamic voxelizer impact on 3D detection metrics on Waymo validation dataset. Latency measured on RTX-3060Ti}
\begin{center}
\scriptsize
\begin{tabular}{lrrrrrrrrr}
\hline
\textbf{Model} & \textbf{Range} & \textbf{mAP/L1} & \textbf{mAPH/L1} & \textbf{mAP/L2} &  \textbf{mAPH/L2} & \textbf{Latency}  \\
\hline\hline
CenterPoint-    & PerType	  & 0.6920 &    0.6716 &	0.6338 &	0.6147 & 56 ms\\
PointPillars    & [0, 30)     &	0.8137 &	0.7936 &	0.7946 &	0.7749 & \\
                & [30, 50)    &	0.6612 &	0.6424 &	0.6098 &	0.5922 & \\
                & [50, +inf)  &	0.5061 &	0.4834 &	0.4235 &	0.4036 & \\
\hline 
RVDet (ours)-   & PerType     & 0.7287 &	0.7095 &	0.6721 &	0.6541 & 43 ms\\
with removed    & [0, 30)     &	0.8266 &	0.8075 &	0.8075 &	0.7888 & \\
ground points   & [30, 50)    &	0.6969 &	0.6775 &	0.6465 &	0.6283 & \\
                & [50, +inf)  &	0.5721 &	0.5530 &	0.4863 &	0.4694 & \\
\hline  
RVDet(ours)     & PerType     &\bf 0.7374 & \bf   0.7176 & \bf 0.6807 & \bf 0.6621 & 47 ms\\
                & [0, 30)     &\bf 0.8323 & \bf   0.8129 & \bf 0.8134 & \bf	0.7945 & \\
                & [30, 50)	  &\bf 0.7065 & \bf   0.6875 & \bf 0.6558 & \bf	0.6379 & \\
                & [50, +inf)  &\bf 0.5840 & \bf   0.5612 & \bf 0.4974 & \bf	0.4772 & \\
\hline
\end{tabular}
\end{center}
\label{tab:dyn_voxel_waymo_val}
\end{table}

\begin{table}[h]
\caption{Dynamic voxelizer detailed metrics on KITTI validation dataset compared with CenterPoint (Diff: difficulty, Mod: Moderate)}
\begin{center}
\scriptsize
\begin{tabular}{l | l | c | c | c | c | c | c | c | c | c | c | c | c  }
\hline
 \multicolumn{2}{c}{ }  & \multicolumn{4}{|c|}{\bf Car} & \multicolumn{4}{c|}{\bf Pedestrian} & \multicolumn{4}{c}{ \bf Cyclist}  \\
\hline
\bf Model &\bf Diff. &\bf BBox &\bf BEV &\bf 3D &\bf AOS &\bf BBox &\bf BEV &\bf 3D &\bf AOS&\bf BBox &\bf BEV & \bf 3D & \bf AOS \\
\hline \hline
CenterPoint-
    & Easy     & 90.62 & 89.66 & 86.60 & 90.61 & 63.45 & 56.07 & 50.84 & 58.76 & 86.83 & 81.26 & 80.21 & 86.56 \\
PointPillars
    & Mod.     & 89.06 & 86.05 & 76.69 & 88.96 & 61.40 & 52.80 & 47.96 & 55.64 & 72.62 & 64.39 & 62.68 & 71.91 \\
    & Hard     & 87.40 & 84.42 & 74.33 & 87.23 & 60.22 & 50.66 & 45.22 & 54.13 & 68.39 & 61.42 & 59.87 & 67.70 \\
\hline
\bf RVDet
    & Easy     &\bf 91.87 & 89.58 &\bf 86.76 &\bf 91.85 &\bf 68.63 & 55.60    &\bf 51.58 &\bf 63.62 &\bf 90.29 &\bf 85.61 &\bf 81.49 &\bf 90.14 \\
\bf (ours)
    & Mod.     & 89.05    & 86.21 & 76.87    & 88.86    &\bf 66.36 &\bf 53.04 &\bf 48.27 &\bf 60.92 &\bf 74.02 &\bf 68.51 &\bf 63.63 &\bf 73.62 \\
    & Hard     &\bf 88.09 &\bf 84.53 &\bf 75.26 &\bf 87.80 &\bf 64.70 & 49.97    & 45.41    &\bf 58.66 &\bf 71.56 &\bf 64.33 &\bf 61.32 &\bf 71.16 \\
\hline
\end{tabular}
\end{center}
\label{tab:dyn_kitti_val}
\end{table}

\section{Ablation Study} \label{sssec:ablation}
To inspect our voxelizer impact on model performance, we validate our trained model on Waymo and KITTI validation sets and report the results in table \ref{tab:dyn_voxel_waymo_val} for Waymo main metrics and in table \ref{tab:dyn_kitti_val} for KITTI mertics.

To choose our classifier architecture, we have carefully run multiple experiments to get the most efficient model in terms of time complexity and accuracy. 
Among the parameters of the classifier architecture that affect the metrics, the following are considered: types of network layers, the number of layers in the network, as well as the size of the layers, depending on the size of the sliced section of the tensor.
To determine the most suitable architecture, experiments were carried out on Waymo Validation set  considering parameters that determine the network architecture. Based on the results of CenterPoint PointPillars detector on Train set, a new dataset was assembled from objects of six classes: True Vehicle, False Vehicle, True Pedestrian, False Pedestrian, True Cyclist, False Cyclist. Each class has 5000 objects. 
Table \ref{tab:mlp_ablation} considers MLP models with different number of layers. The MLP-1-layer model has only 1 linear layer. The size of the input layer of all the linear models listed below is determined by the size of the supplied tensor multiplied by number of channels in the tensor (3 channels in our case). The output of this model is 3 neurons. In the MLP-2-layer model, there are two linear layers, between which the ReLU activation function is applied. At the output of the first layer, the number of neurons is twice as large as input. This number of neurons goes as input size of second layer. There are also 3 neurons at the output of the second layer. In the MLP-3-layer model, the difference is that, only a layer with 24 output neurons and a ReLU activation function is added between the two layers. In the MLP-4-layer model, a layer with 6 output neurons and a ReLU activation function is added before the last layer. In the classifier models with convolutional layers, the size of the MLP input was adjusted to the output of the convolutional layers in front of them. In the 1Conv convolution layer, the parameters: $Number$ $of$ $in$ $channels = 3$, $number$ $of$ $out$ $channels = 6$, $kernel$ $size=(2, 2)$, $stride=(1, 1)$ were used. One more convolutional layer is added to 2Conv with parameters: $Number$ $of$ $in$ $channels = 6$, $number$ $of$ $out$ $channels = 12$, $kernel$ $size=(2, 2)$, $stride=(1, 1)$ 

Table  \ref{tab:mlp_ablation} shows the validation accuracy during Linear-convolutional classifier training on the prepared dataset depending on the number of layers and the size of the input tensor. It can be seen from the tested architectures that, MLP with  8x8 input tensor shows the highest accuracy. For the most detailed study of MLP with such a number of layers, the experiments reflected in Table \ref{tab:best_mlp_ablation} were carried out. In this case, the architecture was examined on a larger number of possible input tensor sizes. We also show Recall of each class for different input sizes.
From the experiments, we conclude that a classifier having two linear layers with 9x9 input tensor  has the best accuracy.
Table \ref{tab:pipeline_ablation} shows experiments demonstrating the effect of the classifier on the precision metric. In this case, we set IoU threshold when calculating the metric to 0.4. To compare the metrics, experiments were carried out with the method of box filtering by the number of LiDAR points that fell inside the predicted box, and an additional filtering of all predictions with a confidence score less than 0.3.

\begin{table}[h]
\caption{Classifier architecture impact on precision of CenterPoint PointPillars on Waymo validation dataset}
\begin{center}
\scriptsize

\begin{tabular}{l | c | c | c | c}
\hline
{\bf Pipeline} & {\bf Overall} & {\bf Vehicle} & {\bf Pedestrian} & {\bf Cyclist}\\ \hline \hline
CenterPoint PP & 31.70 \% & 53.44 \% & 10.77 \% & 4.02  \%\\
CenterPoint PP + Point Filtering (Threshold 5) & 77.48 \% & 90.02 \% & 36.28 \% & 10.04  \%\\
CenterPoint PP + Score Filtering (Threshold 0.3) & 78.91 \% & 90.02 \% & 38.30 \% & 10.38  \%\\
RVCDet (MLP-2-layers) & 85.94 \% & \textbf{94.24} \% & 41.29 \% & \textbf{11.61}  \%\\
RVCDet (2Conv + MLP-2-layers) & \textbf{86.95} \% & 92.56 \% & \textbf{44.39} \% & 11.4  \%\\\hline
\end{tabular}

\end{center}
\label{tab:pipeline_ablation}
\end{table}

\begin{table}[t]
\caption{Accuracy for different classifier architectures on Waymo validation data}
\begin{center}
\scriptsize

\begin{tabular}{l | c | c | c | c | c}
\hline
{\bf Number of layers} & {\bf 2x2 input} & {\bf 4x4 input} & {\bf 6x6 input} & {\bf 8x8 input} & {\bf 10x10 input}\\ \hline \hline
MLP-1-layer  & 89.01 \% & 89.23 \% & 89.61 \% & 89.67 \% & 90.11  \%\\
MLP-2-layers  & \textbf{89.92} \% & 90.08 \% & 90.82 \% & \textbf{90.91} \% & 90.47  \%\\
MLP-3-layers  & 89.48 \% & \textbf{90.29} \% & 90.02 \% & 90.56 \% & 90.31  \%\\
MLP-4-layers  & 89.40 \% & 90.02 \% & 90.40 \% & 90.05 \% & 90.37  \%\\
1Conv+MLP-2-layers & - & - & 89.70 \% & 90.01 \% & 90.43 \%\\
2Conv+MLP-2-layers & - & - & \textbf{91.20} \% & \textbf{90.91} \% & \textbf{90.91} \%\\\hline
\end{tabular}

\end{center}
\label{tab:mlp_ablation}
\end{table}

\begin{table}[t]
\caption{Validation accuracy and recall (for false predictions) for MLP-2-layers with different input sizes on Waymo dataset}
\begin{center}
\scriptsize

\begin{tabular}{l | c | c | c | c }
\hline
{\bf Input} & {\bf Accuracy} & {\bf Recall (Vehicle)}& {\bf Recall (Pedestrian)} & {\bf Recall (Cyclist)}\\ \hline \hline
1x1  & 85.27 \% & 87.22 \% & 84.08 \% & 86.64 \%\\
3x3  & 89.79 \% & 90.75 \% & 86.43 \% & 91.23 \%\\
5x5  & \textbf{91.49} \% & 87.67 \% & \textbf{87.81} \% & 93.04 \%\\
7x7  & 90.91 \% & 91.31 \% & 86.33 \% & 95.12 \%\\
9x9  & 91.43 \% & \textbf{93.29} \% & 87.31 \% & \textbf{95.52} \%\\\hline
\end{tabular}

\end{center}
\label{tab:best_mlp_ablation}
\end{table}

\section{Conclusions}
Achieving real-time performance without affecting the reliability of the network is still a challenge in 3D object detection from LiDAR point clouds.In our work, we addressed this problem and proposed a solution to improve network inference speed and precision at the same time by implementing a fast dynamic voxelizer that works in a $O(N)$ time complexity faster than other current methods. In addition, we presented a lightweight detection sub-head model for classifying predicted objects and filter out false detected objects that significantly improves model precision in a negligible time and computing cost.

%
%
\bibliographystyle{splncs04}
\bibliography{rvcpaper}
\end{document}